\pdfoutput=1

\documentclass[11pt]{article}

\usepackage{ACL2023}

\usepackage{microtype}
\usepackage{times}
\usepackage{latexsym}

\usepackage[T1]{fontenc}

\usepackage[utf8]{inputenc}

\usepackage{inconsolata}

\usepackage{graphicx}
\usepackage{amsmath}
\usepackage{amssymb}
\usepackage{url}
\usepackage{booktabs}
\usepackage{subfigure}
\usepackage{bm}
\usepackage{color}
\usepackage{pifont}
\usepackage{amsfonts}
\usepackage{stmaryrd}
\usepackage{stfloats}
\usepackage{xspace}
\usepackage{threeparttable}
\usepackage{adjustbox}
\usepackage{multirow}
\usepackage{soul}
\usepackage{ragged2e}
\usepackage{enumitem}

\newcommand{\modelname}{MS-DETR\xspace}

\newcommand{\ie}{\emph{i.e.,}\xspace}
\newcommand{\eg}{\emph{e.g.,}\xspace}

\title{\modelname: Natural Language Video Localization with Sampling Moment-Moment Interaction}

\author{
   Jing Wang$^{1,3,4}$, Aixin Sun$^{2}$\thanks{\ \ Corresponding Authors}, Hao Zhang$^{1}$, Xiaoli Li$^{1,3,4}$\footnotemark[1] \\
   $^{1}$ School of Computer Science and Engineering, Nanyang Technological University, Singapore\\
   $^{2}$ S-Lab, Nanyang Technological University, Singapore \\
   $^{3}$ Institute for Infocomm Research, A*STAR, Singapore\\
   $^{4}$ Centre for Frontier AI Research, A*STAR, Singapore\\
   \texttt{\{jing005@e.,axsun@,hao007@e.\}ntu.edu.sg,}~~
   \texttt{xlli@i2r.a-star.edu.sg}
}

\begin{document}
\maketitle

\begin{abstract}
Given a  query, the task of \textit{Natural Language Video Localization} (NLVL) is to localize a temporal moment in an untrimmed video that semantically matches the query. In this paper, we adopt a proposal-based solution that generates proposals (\ie candidate moments)
and then select the best matching proposal. On top of modeling the cross-modal interaction between candidate moments and the query, our proposed Moment Sampling DETR (\modelname) enables efficient moment-moment relation modeling. 
The core idea is to sample a subset of moments guided by the learnable templates with an adopted DETR (DEtection TRansformer) framework. 
To achieve this, we design a multi-scale visual-linguistic encoder, and an anchor-guided moment decoder paired with a set of learnable templates. 
Experimental results on three public datasets 
demonstrate the superior performance of \modelname.\footnote{\small Code is released in \href{https://github.com/K-Nick/MS-DETR}{https://github.com/K-Nick/MS-DETR} }

\end{abstract}
\section{Introduction}
\label{sec:intro}

\begin{figure}[t]
    \centering
    \includegraphics[width=\columnwidth]{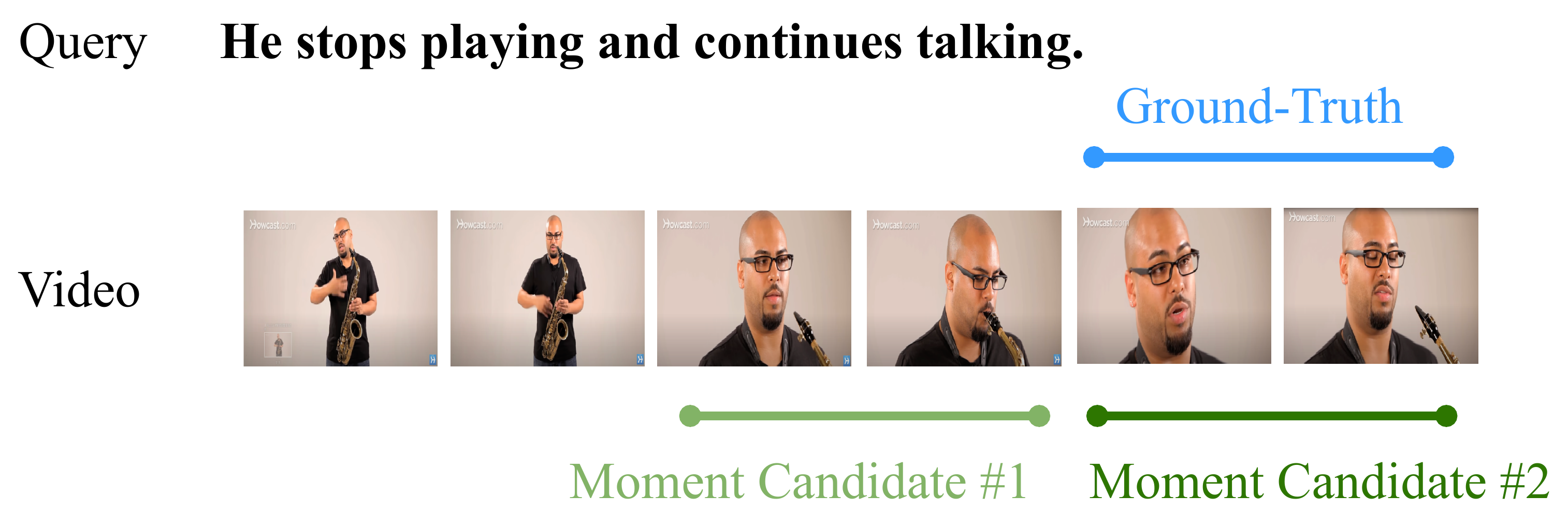}
    \vspace{-8mm}
    \caption{\small An NLVL example with query and ground truth video moment. Two moment candidates with similar video features are also highlighted in light and dark green colors. 
    }
    \label{fig:example}
\end{figure}

Natural language video localization (NLVL) aims to retrieve a temporal moment from an untrimmed video that semantically corresponds to a given language query, see Fig.~\ref{fig:example} for an example. This task is also known as temporal sentence grounding in video, and video moment retrieval. 
As a fundamental video-language task, it has a wide range of applications, such as video question answering~\cite{DBLP:conf/cvpr/FanZZW0H19,DBLP:conf/eccv/YuKK18,DBLP:conf/aaai/LiSGLH0G19}, video retrieval~\cite{DBLP:conf/eccv/Gabeur0AS20,DBLP:conf/bmvc/LiuANZ19,DBLP:conf/cvpr/ChenZJW20}, and video grounded dialogue~\cite{DBLP:conf/acl/LeSCH19, DBLP:conf/aaai/KimYKY21}. 

Generally speaking, in NLVL models, a video is first split to a sequence of many small fixed-length segments. Video features are then extracted from these segments to interact with the text query. Conceptually, each video segment can be viewed as a form of ``video token''.
There are mainly two genres of approaches to NLVL. 
\textit{Proposal-free methods} directly model the interaction between video tokens and text, and aim to identify start/end boundaries along the video token sequence. 
\textit{Proposal-based methods} generate candidate moments as proposals and then select the best matching proposal\footnote{We use the terms \textit{proposal} and \textit{candidate moment} interchangeably, or even simply \textit{moment} when the context is clear.} as the answer. Each proposal is a continuous span of video tokens. 

\begin{figure}[t]
    \centering
    \includegraphics[width=1\columnwidth]{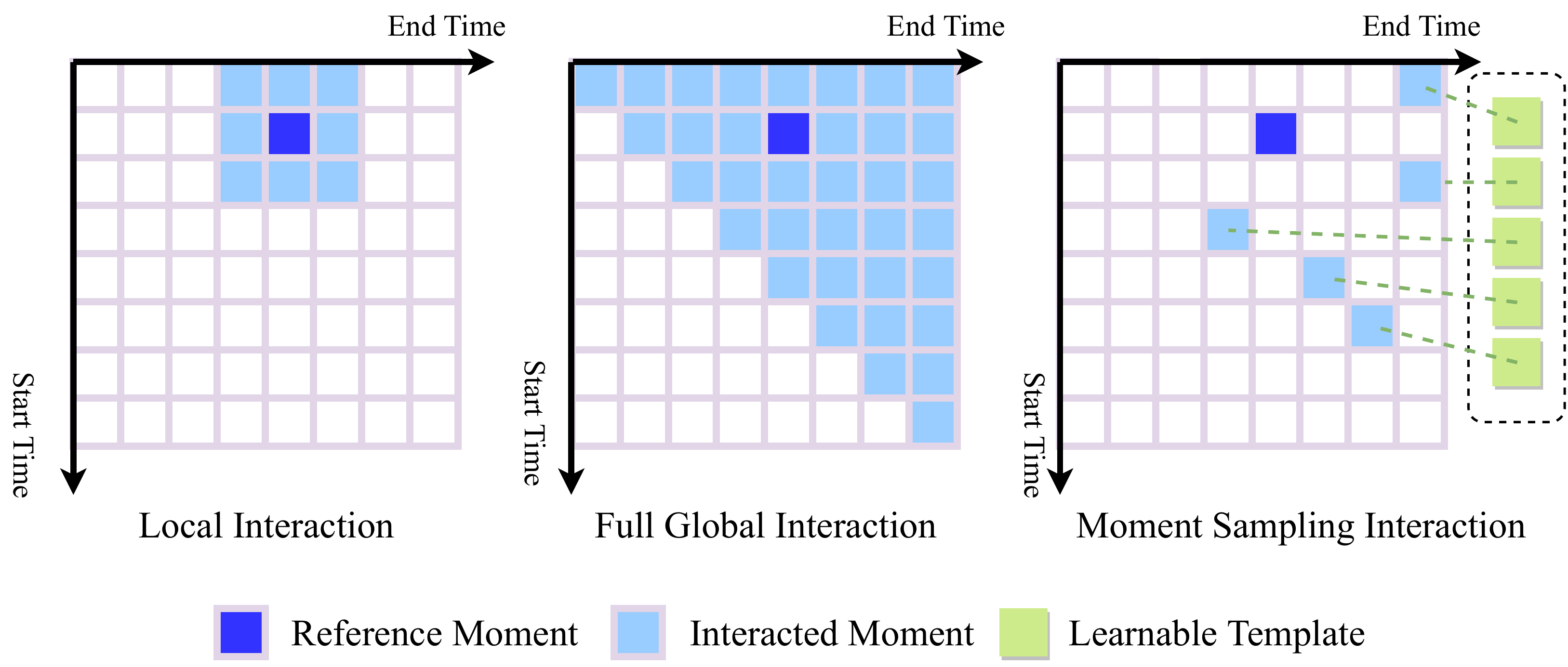}
    \vspace{-6mm}
    \caption{\small Illustration of three strategies of moment-level interactions. Each cell represents a moment with start time $i$ and end time $j$ indicated on the two axes; only the upper triangular area is valid as $i\leq j$.
    }
    \label{fig:strategy}
\end{figure}

To generate proposals, some methods enumerate all possible moment candidates via pre-defined anchors. Anchors are reference start/end positions along the video.  Fig.~\ref{fig:strategy} shows three 2D-Map examples.  Each cell in a 2D-Map corresponds to a candidate moment defined by its start/end time along the two axes. Some other methods produce moment candidates with a proposal generator guided by text query and then refine them independently. The interaction between text and video is mainly modeled between text and video moments; each moment is characterized by the video segments that compose it. Very few studies have considered moment-moment interaction. Consequently, it is challenging to discriminate among moments if there are multiple moments that all demonstrate high level of semantic matching with the text query.
For instance, the two candidate moments in Fig.~\ref{fig:example} have very similar video content and share similar semantic correspondence with the query. 

In this paper, we adopt the proposal-based approach for its capability of cross-modal interaction at both segment level and moment level. We propose \modelname to facilitate effective \textit{text-moment alignment} and efficient \textit{moment-moment interaction}. 
For text-moment alignment, we devise a multi-scale vision-language transformer backbone to conduct segment-word and segment-segment interactions at different segment scales. For moment-moment interaction, our main focus is on which moments should be sampled for interaction, due to the large number of possible pairs. Recall that a moment is a span of segments. Let $\mathcal{O}(N)$ be the magnitude of segment space; the magnitude of moments is $\mathcal{O}(N^2)$. Then moment-moment interaction has a space of $\mathcal{O}(N^4)$.

In practice, not every pair of moments are relevant to each other, and are needed to be discriminated for a given query. Existing methods~\cite{zhangLearning2DTemporal2020,zhangMultiScale2DTemporal2021,wangStructuredMultiLevelInteraction2021} mainly rely on a strong assumption that only the overlapping or adjacent moments are more likely to be relevant, \ie moment locality. An example of moment locality is shown in Fig.~\ref{fig:example}, where two adjacent candidate moments share high level of visual similarity. The local interaction strategy is illustrated in Fig.~\ref{fig:strategy}, where the reference moment only interacts with the surrounding moments in the 2D-Map. 
However, not all relevant moments are overlapping or located close to each other. Following the example in Fig.~\ref{fig:example}, if the person plays saxophone again in the later part of the video (not showing for the sake of space), and the query becomes ``He plays saxophone \textit{again}'', then there will be at least two highly relevant moments for playing saxophone, separated by his action of talking in between. To correctly locate the answer, the model needs to understand that ``\textit{again}'' refers to the second moment of playing saxophone. This calls for a better way of sampling moments for efficient moment-moment interaction, to avoid the full global interaction as shown in Fig.~\ref{fig:strategy}.

The proposed \modelname samples moments for interaction using learnable templates and anchors, illustrated in the third 2D-Map in Fig.~\ref{fig:strategy}.
We design an anchor-guided moment decoder to interact and aggregate moment features from the encoder in an adaptive and progressive manner. A fixed number of learnable templates paired with dynamic anchors are used to match the moment content and its location. Here, the templates are used to match video content in a moment, and anchors specify the reference start/end positions of the moment because multiple moments may share similar visual features. We then revise the anchors based on the predictions from the last decoder block in an iterative manner. We remark that our method has no assumption on moment locality: the moments can be scattered in diverse locations of the video.

Our key contributions are threefold. First,  we propose a novel multi-scale visual-linguistic encoder  (Section~\ref{ssec:mscaleEncoder}) to align textual and video features as well as to aggregate language-enhanced semantics of video frames, in a hierarchical manner. Second, we introduce a new anchor-guided moment decoder (Section~\ref{ssec:momentDecoder}) to decode learnable templates into moment candidates, in which we propose an anchor highlight mechanism to guide the decoding. 
Third, we conduct extensive experiments (Section~\ref{sec:exp}) on three benchmark datasets: ActivityNet Captions, TACoS, and Charades-STA. Our results demonstrate the effectiveness of the proposed \modelname.

\section{Related Work}
\label{sec:related}
We first briefly review existing NLVL approaches and highlight the differences between our work and other proposal-based solutions. Next, we briefly introduce object detection to provide background for the concept of learnable templates. 

\paragraph{Natural Language Video Localization.} NLVL was first introduced in~\citet{DBLP:conf/iccv/HendricksWSSDR17}, and since then a good number of solutions have been proposed~\cite{zhangElementsTemporalSentence2022}.  As aforementioned, existing  methods can be largely grouped into proposal-based and proposal-free methods. 
Proposals, or candidate moments, can be either predefined~\cite{gaoTALLTemporalActivity2017,DBLP:conf/iccv/HendricksWSSDR17} or computed by proposal generator ~\cite{xiaoNaturalLanguageVideo2021,xiaoBoundaryProposalNetwork2021,liuAdaptiveProposalGeneration2021}.
Proposal-free methods output time span~\cite{zhangSpanbasedLocalizingNetwork2020,DBLP:journals/pami/ZhangSJZZG22,zhangParallelAttentionNetwork2021,liuContextawareBiaffineLocalizing2021} or timestamps \cite{yuanFindWhereYou2019, DBLP:conf/naacl/GhoshAPH19,liProposalfreeVideoGrounding2021,zhouEmbracingUncertaintyDecoupling2021} directly on top of video tokens, without considering the notion of candidate moments.

Most proposal-based methods conduct multi-modal interaction between video segments and text, then  encode moments from the segment features. Typically there is no further interactions among moments. 2D-TAN~\cite{zhangLearning2DTemporal2020} is the first to demonstrate the effectiveness of moment-level interaction. However, 2D-TAN assumes moment locality and only enables local interactions among moments as shown in Fig.~\ref{fig:strategy}. However, similar moments requiring careful discrimination may be scattered all over the video. This motivates us to go beyond the moment locality assumption and propose moment sampling for interaction, which is a key difference and also a contribution of our work.

In this paper, we adapt the concept of learnable templates from DETR framework to achieve dynamic moment sampling. DETR was originally introduced for object detection in computer vision (CV), to be briefed shortly. Most similar to our work is~\citet{xiaoNaturalLanguageVideo2021}, which also uses learnable templates. However, their work directly adopts learnable templates without any adaption to the specific requirements of NLVL. For instance, the answer moment in NLVL needs to match the given text query, whereas in object detection, there is no such requirement. We bridge the gap between NLVL and object detection by introducing a hierarchical encoder and a decoder with an anchor highlight mechanism. These designs greatly improve performance and unveil the potential of DETR for NLVL. At the same time, these designs also make our model much different from the original DETR. 

\paragraph{Transformer-based Object Detection.}
Object detection is a fundamental CV task. Transformer-based methods now set a new paradigm that uses learnable templates to sparsely localize objects in  images. The core idea is to aggregate encoder features globally, by using (randomly initialized) learnable templates. To achieve end-to-end detection,  object detection is reformulated as a set prediction problem, \eg certain template combinations can be used to identify some specific image objects. Early solutions match predictions with ground-truth one by one using bipartite matching, leading to unstable matching and slow convergence. Recent work alleviates this issue by designing many-to-one assignment ~\cite{chenGroupDETRFast2022,jiaDETRsHybridMatching2022} or the self-supervision task specifically for learnable templates~\cite{liDNDETRAccelerateDETR2022,zhangDINODETRImproved2022}.

Introducing learnable templates to NLVL poses two challenges: \textit{supervision sparsity} and \textit{scale mismatching}. An image typically contains multiple objects and these co-occurred objects all serve as detection objects for supervision. In NLVL, given a good number of candidate moments in a video, there is only one ground-truth. We refer to this phenomenon as supervision sparsity. The scale extremity in NLVL is more severe than that in object detection. The ground truth moments in videos, analogous to objects in images, vary from 3\% to 90\% in terms of video length.  The diverse scales bring the issue of scale mismatching when the learned templates are decoded to cover all encoder features, \ie the entire video. Hence in \modelname, we adapt learnable templates mainly for the purpose of sparsely sampling moments for interaction, rather than as the main backbone.

\section{Problem Formulation}
We first present how to map video and text into features, and then define NLVL in feature space. 

Let $V=[f_t]_{t=0}^{t=T-1}$ be an untrimmed video with $T$ frames; $L=[w_j]_{j=0}^{j=M-1}$ be a natural language query with $M$ words. 
We uniformly split the video $V$ into $N$ segments (\ie video tokens) and employ a pre-trained video feature extractor to encode these segments into visual features  $\mathbf{V}=[\mathbf{v}_i]_{i=0}^{i=N-1}$. The $M$ words are encoded with pre-trained word embeddings as $\mathbf{L}=[\mathbf{w}_j]_{j=0}^{j=M-1}$. 

Given the video and text query in their encoded features  $(\mathbf{V}, \mathbf{L})$, the task of NLVL is to localize the timestamp pair $(t_s, t_e)$, the start and end timestamp, of the video moment that matches the query. Note that, due to the  uniform split to segments, there is a correspondence between $t_s$ and $t_e$ of the original video and the segment Ids in the segment sequence.

\section{ Method}
\label{sec:method}

\begin{figure*}[t]
    \centering
    \includegraphics[trim={0cm 0.12cm 0cm 0cm},clip,width=0.9\textwidth]{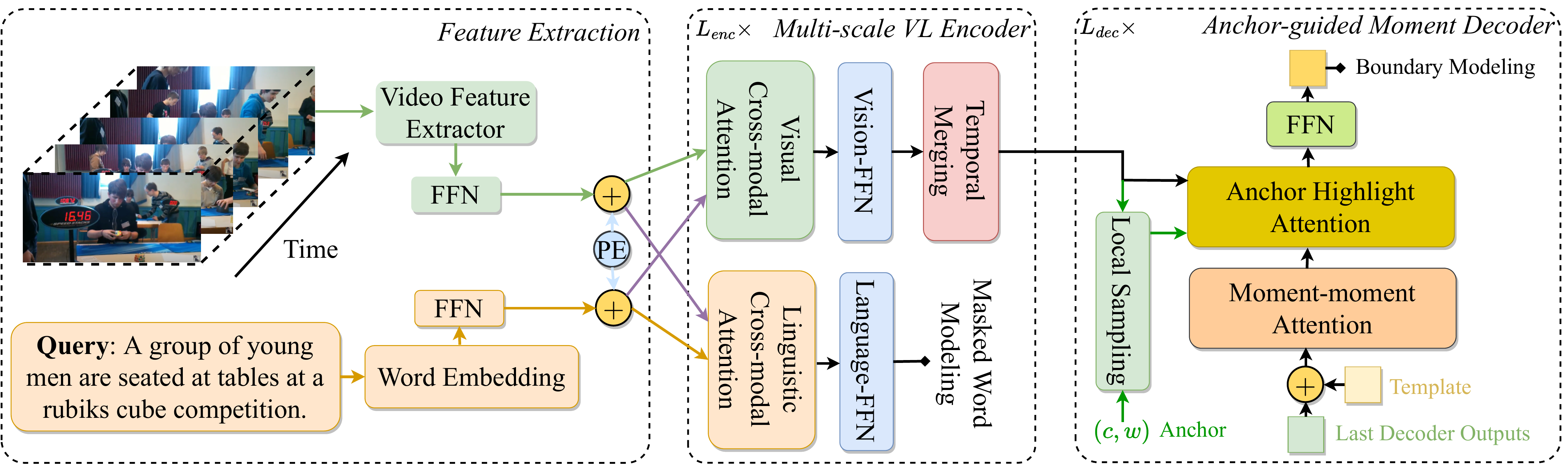}
    \caption{\small The architecture of \modelname for Natural Language Video Localization.}
	\label{fig:architecture}
\end{figure*}

The main architecture of the proposed \modelname is depicted in Fig.~\ref{fig:architecture}. Illustrated in the \textit{feature extraction} part, given visual features $\mathbf{V} \in \mathbb{R}^{d_v \times N}$ and language query features $\mathbf{L} \in \mathbb{R}^{d_w \times M}$, we first project them into a unified dimension $d$ using single layer FFN and decorate them by adding positional encoding, respectively. The linearly projected visual features $\{\mathbf{v}_i^0\}_{i=0}^{i=N-1}$ and language query features $\{\mathbf{w}_j^0\}_{j=0}^{j=M-1}$ are then concatenated and fed into multi-scale vision-language transformer. Next, we mainly detail two main components: multi-scale visual-language encoder, and anchor-guided moment decoder.

\subsection{Multi-scale Visual-Language Encoder}
\label{ssec:mscaleEncoder}

Many transformer-based methods for cross-modal interaction treat video and language tokens identically, in a unified sequence. 
However, video and text have completely different syntactic and semantic structures. It is more reasonable to use separate projections for the two modalities, similar to the idea of modality-specific expert~\citet{pengBEiTV2Masked2022}. 
In \modelname, we separate the projections by using specifically designed attention modules.

Before we further modify the multi-modal attention modules to handle different video resolutions (\ie multi-scale), we present our attention designs in their base form. 
We design two sets of attentions: \textit{visual cross-modal attention} and \textit{linguistic cross-modal attention}, see the middle part of Fig.~\ref{fig:architecture}. The two sets are highly similar. For conciseness, we only introduce visual cross-modal attention, which contains language to video (\textit{L$\rightarrow$V}), and  video to video (\textit{V$\rightarrow$V}) attentions.
The visual cross-modal attention aggregates visual embeddings $\mathbf{V}^l \in \mathcal{R}^{N \times d}$ and language embeddings $\mathbf{L}^l \in \mathcal{R}^{M \times d}$ into new visual features as $\mathbf{V}^{l+1}$: 

\begin{align}
    	\mathbf{A}^{l+1}_{LV} &= \frac{\text{FFN}(\mathbf{V}^l) \text{FFN}(\mathbf{L}^l)}{\sqrt{d_h}} \label{eqn:attenLV} \\
        \mathbf{A}^{l+1}_{VV} &= \frac{\text{FFN}(\mathbf{V}^l) \text{FFN}(\mathbf{V}^l)}{\sqrt{d_h}} \label{eqn:attenVV} \\
        \mathbf{A}^{l+1} &= \mathbf{A}^{l+1}_{LV} \oplus \mathbf{A}^{l+1}_{VV} \label{eqn:attenConcat} \\
        \mathbf{V}^{l+1} &= \text{Softmax}(\mathbf{A}^{l+1}) \notag \\
        &\times\left(\text{FFN}(\mathbf{L}^l)\oplus \text{FFN}(\mathbf{V}^l)\right) \label{eqn:AttenWeightSum} 
\end{align}

The linguistic cross-modal attention uses a similar set of equations to model language to language (\textit{L$\rightarrow$L}) and video to language (\textit{V$\rightarrow$L}) attentions, to get new language features as $\mathbf{L}^{l+1}$.

\paragraph{Sequence-reduced Multi-modal Attention.} Recall that relative lengths of ground truth range from 3\% to 90\% to their source videos. 
A fixed resolution for all moments becomes sub-optimal. 
To this end, we extend the aforementioned multi-modal attention and build a transformer that is capable of providing hierarchical text-enhanced video features, from high to low temporal resolutions. Our encoder design is motivated by the Pyramid Vision Transformer (PVT) \cite{wangPyramidVisionTransformer2021}, which is a successful application of deploying transformer in segmentation problem. 

Handling high temporal resolution is a challenge. Directly applying multi-modal attention on high temporal resolution video features suffers from its quadratic complexity as in Eq.~\ref{eqn:attenVV}.
Recall that the sequence lengths of key, query, and value in multi-head attention \cite{vaswaniAttentionAllYou2017} do not have to be the same. Its output has the same length as the query, and the key-value pair keep the same length. Thus, reducing sequence lengths of the key and value simultaneously is an effective way to save computation. Accordingly, we modify \textit{V$\rightarrow$V} attention in the \textit{visual cross-modal attention} module to a sequence-reduced version as follows:
\begin{align}
    	\mathbf{V}_r^l &= \text{Conv1D}(\mathbf{V}^l) \notag \\
    	\mathbf{A}^{l+1}_{VV} &= \frac{\text{FFN}(\mathbf{V}^l) \text{FFN}(\mathbf{V}^l)}{\sqrt{d_h}} \label{en:attenVV-R}\\
        \mathbf{V}^{l+1} &= \text{Softmax}(\mathbf{A}^{l+1}) \notag\\
        &\times\left(\text{FFN}(\mathbf{L}^l)\oplus \text{FFN}(\mathbf{V}_r^l)\right) 
    \label{eq:seq-reduce}
\end{align}
Here, $\text{Conv1D}$ is a non-overlapping 1D convolution with stride and kernel size set to $R$. 
Eq.~\ref{eqn:attenVV} and Eq.~\ref{eqn:AttenWeightSum} are respectively modified to their new versions in Eq.~\ref{en:attenVV-R} and Eq.~\ref{eq:seq-reduce}. Time complexity is reduced from $\mathcal{O}(N^2)$ to $\mathcal{O}(\frac{N^2}{R})$. We also apply sequence reduction to \textit{V$\rightarrow$L} attention in the \textit{linguistic cross-modal attention}. Conceptually, this sequence reduction technique can be explained as decomposing the local and global interaction. The local interaction is achieved by convolution and the global interaction by attention. Next, we focus on how to merge high to low temporal resolutions. 

\paragraph{Temporal Merging}
To form a hierarchical architecture, a crucial step is a pooling-like step to shrink the temporal scale. We utilize an 1D convolution with overlapping to shrink representations from high to low temporal resolutions. The overlapped convolution allows information flow among convolutional windows, so that the interaction is not constrained locally within windows. With both sequence-reduced multi-modal attention and temporal merging, we form a hierarchical architecture. For the deeper layers in the encoder, which already have a low resolution, we turn off these two components and use the vanilla multi-modal attention.

\paragraph{Auxiliary Supervision Losses} We design two auxiliary losses: \textit{span loss} and \textit{masked word loss}. 
Span loss is to enhance the language-conditioned video representations from encoder. We use the video features $\mathbf{V}^{(L_{enc}-1)}$ from the last layer of encoder to predict whether each video segment falls within the ground truth. This auxiliary supervision facilitates the model to distinguish relevant video segments from irrelevant ones.  We predict span logits $\mathbf{S_{sp}} = FFN(\mathbf{V}^{L_{enc}-1})$ by passing forward encoder output $\mathbf{V}^{(L_{enc}-1)}$ after a two-layer FFN. Span scores $\mathbf{P_{sp}}$ are then calculated from $\mathbf{S_{sp}}$ with a sigmoid function. Then the span loss is computed in Eq.~\ref{eq:span-loss}, where $\mathbf{Y}_{sp}\in\{0, 1\}$.
\begin{equation}
\mathcal{L}_{span} = \left(\log{\mathbf{P}_{sp}} \times \mathbf{Y}_{sp}\right) \times \left(\mathbf{P}_{sp} - \mathbf{Y}_{sp}\right)^2
    \label{eq:span-loss}
\end{equation}
Considering  ground-truth can be a small portion of the source video, 
focal loss \cite{DBLP:journals/pami/LinGGHD20} is adopted here to alleviate the label imbalance issue. 

The masked word loss aims to better align text features and video features. We dynamically replace 15\% words from language query during training with randomly initialized mask embedding. The model is then compelled to learn from both textual and video contexts to recover the missing tokens. Text features $\mathbf{W}^{(L_{enc}-1)}$ from last layer of encoder are used to predict the original words before masking. Masked word score is predicted by  $\mathbf{P_{wm}} = Softmax(\mathbf{S_{wm}})$, where $\mathbf{S_{wm}} = FFN(\mathbf{W}^{(L_{enc}-1)})$. We use the cross entropy loss for masked word prediction.
\begin{equation}
    \mathcal{L}_{mask} = CrossEntropy(\mathbf{P_{wm}}, \mathbf{Y_{wm}})
    \label{eq:word-mask-loss}
\end{equation}

\paragraph{Multi-scale Text-enhanced Features} 
After $L_{enc}$ layers of encoder, we select $C$ text-enhanced video features of different scales from intermediate layer outputs. We re-index the selected outputs $\{\mathbf{V}^{i_0} \cdots \mathbf{V}^{i_{C-1}}\}$ into $\{\mathbf{V}_s^0 \cdots \mathbf{V}_s^{C-1}\}$ for future reference.

\subsection{Anchor-guided Moment Decoder}
\label{ssec:momentDecoder}

After obtaining the multi-scale text-enhanced video features $\mathbf{V}_s=\{\mathbf{V}_s^c\}_{c=0}^{c=C-1}$, our focus now is to decode the learnable templates with their corresponding anchors into moment timestamps. Recall that templates aim to match moment content and anchors are the reference start/end positions. Initially, the anchors are uniformly distributed along the video to guarantee at least one anchor falls within the range of ground truth. 

\begin{figure}[t]
    \centering
    \includegraphics[trim={1.6cm 0.1cm 1.6cm 0.1cm},clip, width=0.8\columnwidth]{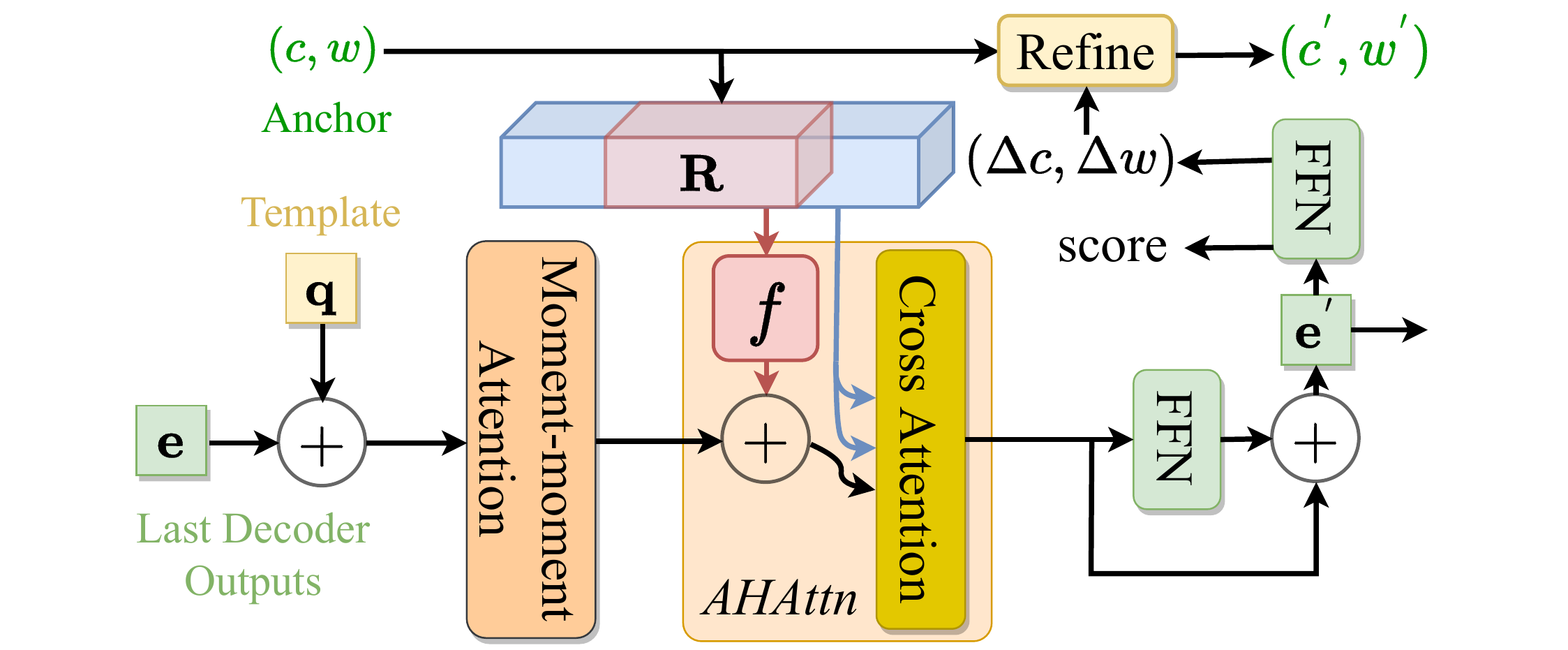}
    \caption{\small Anchor-guided Moment Decoder. Here \textit{AHAttn} is the abbreviation for Anchor Highlight Attention.}
    
	\label{fig:decoder}
\end{figure}

The moment decoder contains two parts: (i) Moment-moment Interaction, which is achieved by self-attention, and (ii) Anchor Highlighting, which aims to not only highlight the area that is relevant to the current moment but also be aware of the global context. The highlighting, or searching for relevant moments, is achieved through an \textit{Anchor Highlight Attention}, an modification of the cross attention in DETR with RoI features, shown in Fig.~\ref{fig:decoder}. 
All attentions mentioned above follow the specification of multi-head scaled dot-product defined in~\citet{vaswaniAttentionAllYou2017}.

\paragraph{Learnable Templates and Anchors}
In the original DETR~\cite{carionEndtoEndObjectDetection2020} paper, the learnable templates can be seen as special positional embeddings, to provide a spatial prior of objects. However, the recent success of advanced DETRs~\cite{liuDABDETRDynamicAnchor2022,mengConditionalDETRFast2021} motivates us to separately model a moment anchor according to which the attention is constrained. Let $k$ denote the index of templates among the total $N_q$ templates. We define $q_k$ as the $k^{th}$ learnable template and $(c^0_k, w^0_k)$ as its initial anchor. Here $c$ and $w$ stand for the center and width of the corresponding moment, which can be easily mapped to the start/end boundary.
Anchors will be refined in the decoder, layer by layer. We use $(c_k^l, w_k^l)$ to denote the anchor after refinement of the $l^{th}$ decoder layer. 

\begin{figure}[t]
    \centering
    \includegraphics[clip, width=0.75\columnwidth]{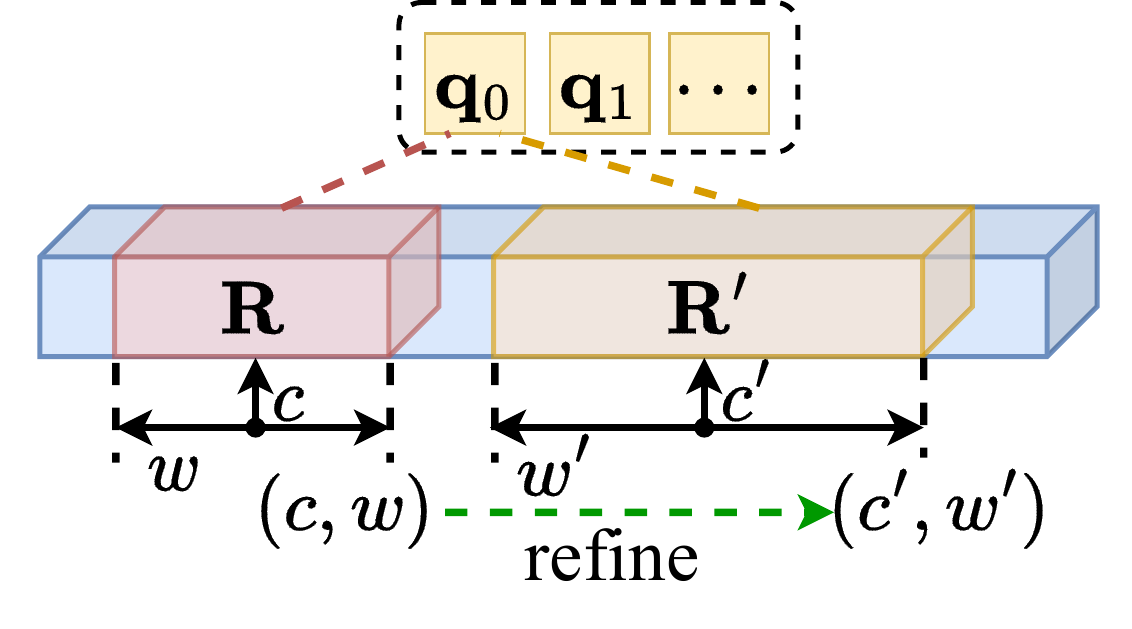}
    \caption{\small An example of anchor refinement. The anchor $(c,w)$ paired with learnable template $q_0$ is refined to $(c',w')$. Accordingly, its moment contents shift from $\mathbf{R}$ to $\mathbf{R}'$.
    }
    \label{fig:refine}
\end{figure}

\paragraph{Anchor Highlight Attention.} One of our motivations is to discriminate the best matching moment among all candidate moments that share good matching to the text query.
To highlight the areas that are similar to the current moment, we modify the attention query to adjust attention weight.

Suppose the current anchor is $(c_k, w_k)$, we can easily locate the corresponding area in the $n^{th}$ multi-scale feature from the encoder output. We use  $\mathbf{r}_{c,k}$ to denote the features in this area that are taken from the $c^{th}$ multi-scale video features $\mathbf{V}_s^c$. Let $\mathbf{R}_k$ be the collection of features from all scales. We then construct a function $\mathbf{f}$ to map  $\mathbf{R}_k$ to a single vector $\mathbf{H}_k \in \mathbb{R}^{d}$ to guide the highlight in attention mechanism, illustrated in Fig.~\ref{fig:decoder}. Let $\mathbf{H}^{N_q \times d} \in \mathbb{R}^{}$ be the collection of $\mathbf{H}_k$ and $\mathbf{M}$ be the moment features after self-attention module in decoder layer, we adjust the attention as follows: 
 \begin{equation}
 	\begin{aligned}
 		\mathbf{A}_{AH} &= \frac{FFN(\mathbf{M} + \mathbf{H}) FFN(\mathbf{V}^{C-1}_s)^\mathsf{T}}{\sqrt{d_h}} \\
        \mathbf{M}_{AH} &= \mathbf{A}_{AH} \times FFN(\mathbf{V}^{C-1}_s)
 	\end{aligned}
    \label{eq:attentionHighlight}
\end{equation}
Here, $A_{AH}$ refers to the adjusted attention weight, and $\mathbf{M}_{AH}$ is the output of the adjusted cross attention. Since $\mathbf{H}$ is sampled and transformed from the corresponding anchor areas in encoder outputs $\mathbf{V}_s$, it is essentially the representation of moment content. Therefore, the term $\mathbf{H}(\mathbf{V}_s^{C-1})^T$ will be more responsive when a specific area from $\mathbf{V}_s^{C-1}$ is similar to the moment content. Consequently, the attention above will highlight the areas similar to the current moment. We then refine the anchors based on these highlighted areas, through an offset prediction head as shown in Fig.~\ref{fig:decoder}.

\paragraph{Anchor Refinement.} Based on the predictions from the last decoder block, we revise the anchor with the predicted offsets.  This is analogous to the eye skimming process of humans: focuses on a local area in the video and then decides where to move her sight at the next step. The anchors are refined iteratively as shown in Fig.~\ref{fig:refine}.  Specifically, we first project the center $c_k^l$ and scale $s_k^l$ of the $k^{th}$ anchor at the $l^{th}$ decoder level into logit space, using an inverse sigmoid operation. The offset $(\Delta c_k^l, \Delta w_k^l)$ is added to their logits, then the modified logits are projected back using sigmoid. The whole process is described in Eq.~\ref{eq:anchor-refine}.
\begin{equation}
    \begin{aligned}
    	c_k^{m+1} &= \sigma\left(\Delta c_k^m + \sigma^{-1}(c_k^m)\right) \\
        w_k^{m+1} &= \sigma\left(\Delta w_k^m + \sigma^{-1}(w_k^m)\right)
    \end{aligned}
    \label{eq:anchor-refine}
\end{equation}
Here $\sigma$ stands for sigmoid function, and $\sigma^{-1}$ for inverse sigmoid function. 

\paragraph{Boundary Modeling.} After encoding moment candidate features, we pass them through two separate FFNs to predict anchor offset and scores, respectively. Depending on anchor positions, only a small portion of anchors may match with ground truth moments. Among them, we simply select the candidate moment with the largest IoU (intersection over union) with ground truth as our positive sample. A similar label assignment strategy has been used in early studies \cite{carionEndtoEndObjectDetection2020}.

After labeling predictions as positive or negative, we refer to the index of positive prediction as $i_p$. Then we model the boundary with two losses: (i) IoU prediction loss, and (ii) $L1$ regression loss. Note that,  $L1$ regression loss is only applied to the positive prediction. Let $(t_s^k, t_e^k)$ be the timestamps predicted by $k^{th}$ anchor and $(t_s^{g}, t_e^{g})$ be the ground-truth timestamps, we calculate $L1$ regression loss and IoU prediction loss as follows:
\begin{equation}
    \begin{aligned}
    	\mathcal{L}_{L1} &= \frac{1}{2} \left( \lvert t_s^{i_p} - t_s^{g} \rvert + \lvert t_e^{i_p} - t_e^{g} \rvert \right) \\
        \mathcal{L}_{IoU} &= \frac{1}{N_q} \sum_{k\in N_q} \text{Focal}(\text{TrIoU}_k, o_k)
    \end{aligned}
    \label{eq:boundary-modeling}
\end{equation}
Here $TrIoU$ truncates IoU between $(t_s^k, t_e^k)$ and $(t_s^g,t_e^g)$ below a threshold $\theta$ and set IoU of the assigned positive prediction to 1. Different from \citet{carionEndtoEndObjectDetection2020}, by using $TrIoU$, we not only calculate IoU loss for the positive prediction but also consider the hard negative predictions which have large overlapping with ground-truth. Note that, IoU prediction loss and $L1$ regression loss are calculated for all decoder layer outputs.

\subsection{Training and Inference}

The overall training loss of \modelname contains three losses: 
\begin{equation}
	\begin{aligned}
      \mathcal{L} &= \lambda_{span} \mathcal{L}_{span} + \lambda_{mask} \mathcal{L}_{mask} \\
      &+\sum_{m\in L_{dec}}\left(\lambda_{IoU} \mathcal{L}_{IoU} + \lambda_{L1} \mathcal{L}_{L1}\right)
	\end{aligned}
    \label{eq:sum-loss}
\end{equation}
To stabilize training, we introduce an extra denoising group of templates and pass them through the decoder,  motivated by \cite{chenGroupDETRFast2022}. The overall loss is averaged over losses calculated from two groups independently.
During inference, we deprecate the denoising group and use the main group only. All moments are sorted by their scores and their anchors are converted from $(c,w)$ to start/end format. We apply truncation to start/end timestamps to deal with out-of-range values, since no constraint is attached to $(c,w)$ during training.
\section{Experiments}
\label{sec:exp}

We evaluate \modelname against baselines on three public benchmarks: ActivityNet Captions \cite{DBLP:conf/iccv/KrishnaHRFN17},  TACoS \cite{DBLP:journals/tacl/RegneriRWTSP13}, and Charades-STA \cite{gaoTALLTemporalActivity2017}. The three datasets cover videos from different domains and lengths (see Appendix~\ref{ssec:appen-dataset} for video distributions and train/dev/test splits). 

Following prior work \cite{zhangParallelAttentionNetwork2021}, we adopt ``$R@n, IoU=\mu$'' and ``mIoU'' as evaluation metrics. $R@n, IoU=\mu$ is the percentage of testing samples that have at least one of top-$n$ results hitting ground truth, where ``hitting'' means an overlapping with $IoU \geq \mu$. mIoU denotes the average IoU over all test samples. We set $n=1$ and $\mu=\{0.3, 0.5, 0.7\}$. In our comparison and discussion, we mainly focus on $\mu=0.7$ as large IoU means high-quality matching.

\subsection{Comparison with the State-of-the-Arts}
Results on the three datasets are compared in Tables~\ref{tab:sota-anet},~\ref{tab:sota-tacos}, and~\ref{tab:sota-charades}, respectively. Baseline results are mostly cited from~\cite{zhangParallelAttentionNetwork2021}. We also include  GTR~\cite{caoPursuitDesigningMultimodal2021}, LP-Net~\cite{xiaoNaturalLanguageVideo2021} and MMN~\cite{wangNegativeSampleMatters2022} for a complete comparison.

\modelname achieves the best $R@1, \mu=0.7$ and mIoU on ActivityNet and TACos, and the second best on Charades-STA. Our model achieves reasonably good results on smaller $\mu$'s. However, large $\mu$ ensures high-quality matching. A possible reason for the results on Charades-STA is that the videos in this dataset are very short (30 seconds on average), making moment-level interaction less necessary.

\begin{table}[t]
    \small
    \centering
	\setlength{\tabcolsep}{7.0 pt}
	\begin{tabular}{l | c c c | c}
		\toprule
		\multirow{2}{*}{Method} & \multicolumn{3}{c |}{$\text{R@}1, \text{IoU}=\mu$} & \multirow{2}{*}{mIoU} \\
        & $\mu=0.3$ & $\mu=0.5$ & $\mu=0.7$ & \\
        \midrule
            DEBUG   & 55.91 & 39.72 & -     & 39.51 \\
            ExCL    & 63.00 & 43.60 & 24.10 & -     \\
            SCDM    & 54.80 & 36.75 & 19.86 & -     \\
            CBP     & 54.30 & 35.76 & 17.80 & -     \\
            GDP     & 56.17 & 39.27 & -     & 39.80 \\
            2D-TAN  & 59.45 & 44.51 & 27.38 & -     \\
            TSP-PRL & 56.08 & 38.76 & -     & 39.21 \\
            TMLGA   & 51.28 & 33.04 & 19.26 & -     \\
            VSLNet  & \underline{63.16} & 43.22 & 26.16 & 43.19 \\
            DRN     & -     & 45.45 & 24.36 & -     \\
            LGI     & 58.52 & 41.51 & 23.07 & -     \\
            SeqPAN  & 61.65 & 45.50 & 28.37 & \underline{45.11} \\
            GTR              & -        & \textbf{49.67}     & 28.45     & -     \\
            LP-Net            & \textbf{64.29}    & 45.92     & 25.39     & 44.72 \\
            MMN             &   65.05     &     48.59   &   \underline{29.26}   &   - \\
            \midrule
            \modelname       & 62.12     & \underline{48.69}     & \textbf{31.15}     & \textbf{46.82} \\
        \bottomrule
	\end{tabular}
	\caption{\small Results on ActivityNet Captions. The best results are in bold face and second best underlined.}
	\label{tab:sota-anet}
\end{table}

\begin{table}[t]
    \small
    \centering
	\setlength{\tabcolsep}{7.0 pt}
	\begin{tabular}{l | c c c | c}
		\toprule
		\multirow{2}{*}{Method} & \multicolumn{3}{c |}{$\text{R@}1, \text{IoU}=\mu$} & \multirow{2}{*}{mIoU} \\
        & $\mu=0.3$ & $\mu=0.5$ & $\mu=0.7$ & \\
        \midrule
				TGN    		& 21.77 & 18.90 & -     & -     \\
                ACL    		& 24.17 & 20.01 & -     & -     \\
                DEBUG  	& 23.45 & 11.72 & -     & 16.03 \\
                SCDM	& 26.11 & 21.17 & -     & -     \\
                CBP    		& 27.31 & 24.79 & 19.10 & 21.59 \\
                GDP    		& 24.14 & 13.90 & -     & 16.18 \\
                TMLGA  & 24.54 & 21.65 & 16.46 & -     \\
                VSLNet 	& 29.61 & 24.27 & 20.03 & 24.11 \\
                DRN    		& -     & 23.17 & -     & -     \\
                SeqPAN 		& 31.72 & 27.19 & \underline{21.65} & \underline{25.86} \\
                DRN					&	-	& 23.17	&	-	&	-	\\
                CMIN			& 24.64   & 18.05   & -       & -   \\
                2D-TAN            & 37.29   & 25.32   & -       & -   \\
                GTR              	& \underline{40.39}   & \underline{30.22}   & -       & -   \\
                MMN             &   39.24     &     26.17   &   -   &   - \\
                \midrule
                \modelname       & \textbf{47.66}   & \textbf{37.36}   &\textbf{25.81} & \textbf{35.09} \\
        \bottomrule
	\end{tabular}
	\caption{\small Results on TACoS, best results in bold face, and second best underlined.}
	\label{tab:sota-tacos}
\end{table}

\begin{table}[t]
    \small
    \centering
	\setlength{\tabcolsep}{7.0 pt}
	\begin{tabular}{l | c c c | c}
		\toprule
		\multirow{2}{*}{Method} & \multicolumn{3}{c |}{$\text{R@}1, \text{IoU}=\mu$} & \multirow{2}{*}{mIoU} \\
        & $\mu=0.3$ & $\mu=0.5$ & $\mu=0.7$ & \\
        \midrule
            DEBUG            & 54.95   & 37.39   & 17.69   & 36.34 \\
            ExCL             & 61.50   & 44.10   & 22.40   & -     \\
            MAN              & -       & 46.53   & 22.72   & -     \\
            SCDM             & -       & 54.44   & 33.43   & -     \\
            CBP              & -       & 36.80   & 18.87   & -     \\
            GDP              & 54.54   & 39.47   & 18.49   & -     \\
            2D-TAN           & -       & 39.81   & 23.31   & -     \\
            TSP-PRL          & -       & 45.30   & 24.73   & 40.93 \\
            MMN			&	47.31 & 27.28	&	-	&	- \\
            VSLNet           & 70.46   & 54.19   & 35.22   & 50.02 \\
            LGI              & \underline{72.96}   & \underline{59.46}   & 35.48   & -     \\
            SeqPAN           & \textbf{73.84}   & \textbf{60.86}   & \textbf{41.34}   & \textbf{53.92} \\
            \midrule
            \modelname       & 68.68   & 57.72   & \underline{37.40}   & \underline{50.12} \\
        \bottomrule
	\end{tabular}
	\caption{\small Results on Charades-STA, best results in bold face, and second best underlined.}
	\label{tab:sota-charades}
\end{table}

\subsection{Ablation Study}

We perform ablation studies on ActivityNet Captions for the effectiveness  \modelname. 

\paragraph{Multi-scale Encoder.}
We evaluate four variants to study the effectiveness of multi-scale design in our transformer encoder. First, to evaluate whether hierarchical design benefits cross-modal interaction, the `uni-scale' variant replaces all sequence-reduced layers with normal layers without resolution shrinkage, and set the number of clips to 32. The multi-scale transformer now degrades to a uni-scale cross-modal transformer. 
To study the contribution of encoding moment contents $R$ for anchor highlighting in multiple scales, the `single-scale' variant  selects the output of the last encoder layer only and fuses it to attention query, while keeping encoder's hierarchical structure.
Then, we study the effect of arranging sequence-reduced layers in different positions in the 5 encoder layers. We compare two arrangements ``BBBRR'' and ``RBBBR'' against \modelname's ``RRBBB''. Here `R' means sequence-reduced and `B' means the base version. 

Results in Table~\ref{tab:ms-ablation} suggest the effectiveness of multi-scale hierarchical encoder. Performance drops with the removal of the multi-scale mechanism, or the other arrangement of sequence-reduced layers.
Placing sequence-reduced version at shallow layers serves the purpose of reducing computational cost while benefiting performance.

\begin{table}[t]
    \small
    \centering
	\begin{tabular}{l | c c c | c}
		\toprule
		\multirow{2}{*}{Method} & \multicolumn{3}{c |}{$\text{R@}1, \text{IoU}=\mu$} & \multirow{2}{*}{mIoU} \\
        & $\mu=0.3$ & $\mu=0.5$ & $\mu=0.7$ & \\
        \midrule
        \modelname          &   \textbf{62.12}   &   \textbf{48.29}   & \textbf{31.15}     & \textbf{46.82} \\      
        uni-scale            &   61.08   &   47.85   &  30.69   &  45.62     \\
        single-scale   &  61.57   &   47.86   &  30.91   &  45.86     \\
        BBBRR &   60.99   &   46.97   &   30.00   & 44.84 \\
        RBBBR &   61.42   &   47.14   &   30.05   & 45.48 \\
        \bottomrule
	\end{tabular}
	\caption{\small Ablation study on multi-scale hierarchical encoder.}
	\label{tab:ms-ablation}
\end{table}

\paragraph{Anchor Highlight Attention} is a variant of standard cross attention~\citet{vaswaniAttentionAllYou2017}. It is used to highlight similar content with corresponding moments across the video. We compare its design with the standard cross attention. Table~\ref{tab:ah-ablation} shows that anchor highlight attention outperforms standard cross attention, by a large margin. This justifies the advantage of using anchor highlight attention and dynamic anchor jointly, to narrow the range of attention to anchor areas.

\begin{table}[t]
    \small
    \centering
	\begin{tabular}{l | c c c | c}
		\toprule
		\multirow{2}{*}{Methods} & \multicolumn{3}{c |}{$\text{R@}1, \text{IoU}=\mu$} & \multirow{2}{*}{mIoU} \\
        & $\mu=0.3$ & $\mu=0.5$ & $\mu=0.7$ & \\
        \midrule
        \modelname          &   \textbf{62.12}   &   \textbf{48.29}   & \textbf{31.15}     & \textbf{46.82} \\      
        CrossAtten.   &   61.25   &   46.05   &   27.94   & 44.30     \\
        \bottomrule
	\end{tabular}
	\caption{\small Anchor highlight attention versus standard cross attention without anchor highlighting.}
	\label{tab:ah-ablation}
\end{table}

\paragraph{The Auxiliary Loss.}
We use two auxiliary supervision losses, span loss and word mask loss, in our encoder (see Section~\ref{ssec:mscaleEncoder}). Table~\ref{tab:aux-loss-ablation} reports the results of removing either one or both auxiliary losses. Results suggest that both auxiliary losses benefit \modelname, and span loss contributes more to the effectiveness of \modelname. That is, supervising encoder to discriminate whether segments fall within the ground-truth area is important for vision-language alignment. 

\begin{table}[t]
    \small
    \centering
	\begin{tabular}{l | c c c | c}
		\toprule
		\multirow{2}{*}{Methods} & \multicolumn{3}{c |}{$\text{R@}1, \text{IoU}=\mu$} & \multirow{2}{*}{mIoU} \\
        & $\mu=0.3$ & $\mu=0.5$ & $\mu=0.7$ & \\
        \midrule
        \modelname          &   \textbf{62.12}   &   \textbf{48.29}   & \textbf{31.15}     & \textbf{46.82} \\   
        w/o $\mathcal{L}_{span}$    &  58.67    &   45.75   &   30.15   &   44.06   \\
        w/o $\mathcal{L}_{mask}$    &  62.04    &   47.9    &  30.17     &   45.40  \\
        w/o both                    &  57.50   &   46.07   &   30.03   &   43.82   \\
        \bottomrule
	\end{tabular}
	\caption{\small The impact of auxiliary losses. }
	\label{tab:aux-loss-ablation}
\end{table}

\paragraph{Hyper-parameter Study.} Results of the choices of the number of encoder/decoder blocks, and number of denoising groups for training stabilization are in Appendix~\ref{ssec:appen-hyper}.

\section{Conclusion}
In this paper, we adapt DETR framework from object detection to NLVL. With the proposed MS-DETR, we are able to model  moment-moment interaction in a dynamic manner. Specifically, we design a multi-scale visual-linguistic encoder to learn hierarchical text-enhanced video features, and an anchor-guided moment decoder to guide the attention with dynamic anchors for iterative anchor refinement. The promising results on three benchmarks suggest that moment-moment interaction for NLVL can be achieved in an efficient and effective manner.

\newpage
\section{Limitation}
The limitation of this paper are twofold. First, our method does not provide a recipe for data imbalancement in  NLVL task. Thus, our method does not guarantee the effectiveness on edge cases.
Second, the choice of feature extractor is considered relatively outdated. Our model does not benefit from the recent development of pre-trained vision-language models. On the other hand, using pre-trained vision-language models remains in its early stage in NLVL tasks. Not using pre-trained features makes a fair comparison between our model with existing baselines. As a part of future work, we will explore the potential of using more powerful feature extractors in our model.
\section{Acknowledgement}
This study is supported under the RIE2020 Industry Alignment Fund – Industry Collaboration Projects (IAF-ICP) Funding Initiative, as well as cash and in-kind contribution from the industry partner(s).
\bibliography{anthology, custom}
\bibliographystyle{acl_natbib}

\appendix
\newpage
\section{Appendix}
\label{Sec:appendix}

\subsection{Dataset Details}
\label{ssec:appen-dataset}
\textbf{ActivityNet Captions} \cite{DBLP:conf/iccv/KrishnaHRFN17} contains over 20K videos paired with 100K queries with an average duration of 2 minutes. We use the dataset split ``val\_1'' as our validation set and ``val\_2'' as our testing set. In our setting, $37,417$, $17,505$, and $17,031$ moment-sentence pairs are used for training, validation, and testing, respectively. \textbf{TACoS} \cite{DBLP:journals/tacl/RegneriRWTSP13} includes 127 videos about cooking activities. The average duration of videos in TACoS is 7 minutes.
We follow the standard split which includes $10,146$, $4,589$, and $4,083$ moment-sentence pairs for training, validation, and testing. \textbf{Charades-STA} \cite{gaoTALLTemporalActivity2017} is built on Charades and contains $6,672$ videos of daily indoor activities. Charades-STA has $16,128$ sentence-moment pairs in total, where $12,408$ pairs are for training and $3,720$ pairs for testing. The average duration of the videos is 30s.

\subsection{Implementation Details}
\label{ssec:appen-implemnetation}

We use AdamW with learning rate of $3\times10^{-4}$ and batch size of 32 for optimization. We follow \cite{zhangLearning2DTemporal2020} and use pretrained 3D Inception network to extract features from videos.  The number of sampled video frames is set to 512 for ActivityNet Caption and TACoS and 1024 for Charades-STA. For \modelname architecture, we use a 5-layers encoder and a 5-layers decoder with all hidden sizes set to 512. For inference, we select the proposal with highest score from the last decoder layer as our prediction. As for the specific choice of $f$ mentioned in Section~\ref{ssec:momentDecoder}, we use RoIAlign to extract multi-scale feature $R$, then concatenate them and pass them through an FFN. One extra denoising group is used for stabilizing training. The loss is then averaged over two groups.  During inference, No extra operation like Non-Maximum Suppression (NMS) is required. All experiments are run on a single A100 GPU. The reported versions take roughly 8-10 GPU hours for training.

\begin{table}[t]
    \small
    \centering
	\begin{tabular}{l | c c c | c}
		\toprule
		\multirow{2}{*}{Methods} & \multicolumn{3}{c |}{$\text{R@}1, \text{IoU}=\mu$} & \multirow{2}{*}{mIoU} \\
        & $\mu=0.3$ & $\mu=0.5$ & $\mu=0.7$ & \\
        \midrule
        \modelname-Enc3     &   62.47   &   48.15   &   30.54   &   45.8   \\
        \modelname-Enc4     &   61.17   &   47.87   &   30.41   &   44.9    \\
        \modelname          	&   \textbf{62.12}   &   \textbf{48.29}   & \textbf{31.15}     & \textbf{46.82} \\      
        \modelname-Enc6     &   62.05   &   48.00   &   31.03   &   45.71   \\
        \bottomrule
	\end{tabular}
	\caption{\small The impact on number of encoder layers. }
	\label{tab:enc-ablation}
\end{table}

\begin{table}[t]
    \small
    \centering
	\begin{tabular}{l | c c c | c}
		\toprule
		\multirow{2}{*}{Methods} & \multicolumn{3}{c |}{$\text{R@}1, \text{IoU}=\mu$} & \multirow{2}{*}{mIoU} \\
        & $\mu=0.3$ & $\mu=0.5$ & $\mu=0.7$ & \\
        \midrule
        \modelname-Dec2     &  60.53   &   46.54   &   29.12   &   44.4   \\
        \modelname-Dec3     &   \underline{62.42}   &   \underline{47.92}   &   30.11   &   45.4   \\
        \modelname-Dec4     &   61.14   &   47.03   &   30.13   &   44.7    \\
        \modelname          &   \textbf{62.12}   &   \textbf{48.29}   & \underline{31.15}     & \textbf{46.82} \\      
        \modelname-Dec6     &   61.3   &   47.83   &   \textbf{31.65}   &   \underline{46.3}   \\
       
        \bottomrule
	\end{tabular}
	\caption{\small The impact of the number of decoder layers. }
	\label{tab:dec-ablation}
\end{table}

\begin{table}[t]
    \small
    \centering
	\begin{tabular}{l | c c c | c}
		\toprule
		\multirow{2}{*}{Methods} & \multicolumn{3}{c |}{$\text{R@}1, \text{IoU}=\mu$} & \multirow{2}{*}{mIoU} \\
        & $\mu=0.3$ & $\mu=0.5$ & $\mu=0.7$ & \\
        \midrule
        \modelname-DN0     &   61.50   &   47.94   &   30.83   &   45.04    \\
        \modelname          &   \textbf{62.12}   &   \textbf{48.29}   & \textbf{31.15}     & \textbf{46.82} \\      
        \modelname-DN2     &   62.13   &   47.74   &   30.91   & 45.6     \\
        \modelname-DN3     &   62.03   &   47.55   &   30.84   & 45.5   \\
        
        \bottomrule
	\end{tabular}
	\caption{\small The impact of the number of denoising groups.}
	\label{tab:dn-ablation}
\end{table}

\subsection{Hyper-parameter Study}
\label{ssec:appen-hyper}

\paragraph{Number of Encoder/Decoder Blocks}

We study the impact of the number of encoder and decoder blocks. We evaluate one of them from 2 to 6, while keeping the other fixed to 5.
The performance across various numbers of encoder and decoder blocks are listed in Table~\ref{tab:enc-ablation}. Best performance is achieved by $L_{enc}=5$ and $L_{dec}=5$. Though the setting of $L_{dec}=6$ has slightly larger ``$R@1,IoU=0.7$'', poorer ``$mIoU$'' is observed. We speculate that the cause is overfitting on some overly confident examples.

\paragraph{Number of Denoising Groups.} We study the effectiveness of using different numbers of denoising groups in training stabilization. The results are evaluated with the number of denoising groups ranging from 0 to 3, in Table~\ref{tab:dn-ablation}. We observe the performance increases after using one denoising group, then gradually decreases. We suspect there is a trade-off between training stability and the ability to escape local minima.

\end{document}


\maketitle

\appendix

\noindent This appendix contains two sections. Section~\ref{apd:sec:model} provides (\ref{apd:sec:sgpa}) a detailed comparison between the proposed SGPA and standard transformer blocks, (\ref{apd:sec:vqi})  technical details of the video-query integration module,  and (\ref{apd:sec:gumbel}) categorical reparameterization used in the sequence matching module. Section~\ref{apd:sec:settings} describes statistics on the benchmark datasets and parameter settings in our experiments.

\section{Additional Comparison and Technical Details}\label{apd:sec:model}

\subsection{SGPA versus Standard Transformers}\label{apd:sec:sgpa}
Two ways are mainly used to adopt the transformer block for multi-modal representation learning: 
\begin{itemize}[itemsep=-0.3ex,leftmargin=2.5ex]
    \item Transformer block with the self-attention (Se-TRM), which encodes visual and textual inputs in separate streams, shown in Figure~\ref{fig:se_trm}.
    \item Transformer block with the cross-attention (Co-TRM), which encodes both visual and textual inputs with interactions through co-attention, shown in  Figure~\ref{fig:co_trm}.
\end{itemize}

Several works~\cite{lu2019debug,chen2020rethinking,zhang2020vslnet} adopt Se-TRM to learn visual and textual representations in video grounding task. Se-TRM separately encodes each modality, it focuses on learning the refined unimodal representations within each modality for video and text respectively. Without any connection between two modalities, Se-TRM cannot use information from other modality to improve the representations. 

Co-TRM\footnote{It is also known as co-attentional, multi-modal or cross-modal transformer block in different works.} is commonly used as a basic component in various vision-language methods~\cite{tan2019lxmert,lu2019vilbert,lei2020tvr}. Co-TRM relies on co-attention to learn the cross-modal representations for both visual and textual inputs. However, Co-TRM lacks the ability to encode self-attentive context within each modality.

The cascade of Se-TRM and Co-TRM is also used in recent vision-language models~\cite{tan2019lxmert,lu2019vilbert,zhu2020actbert,lei2020tvr} to learn both unimodal and cross-modal representations. In general, there are two cascade forms: 1) stacking Co-TRM upon Se-TRM (SeCo-TRM) in Figure~\ref{fig:seco_trm}; and 2) stacking Se-TRM upon Co-TRM (CoSe-TRM) in Figure~\ref{fig:cose_trm}. These stacked TRMs learn the unimodal and cross-modal information in a sequence manner. Hence, their final outputs focus more on either the self-attentive contexts or cross-modal interactions. 
Our SGPA combines advantages of both Se-TRM and Co-TRM, but not through cascade. As shown in Figure~\ref{fig:sgpa_trm}, SGPA contains two parallel multi-head attention blocks. One block takes single modality as input and the other takes both modalities as inputs. Thus, SGPA is able to learn both unimodal and cross-modal representations simultaneously. Then, a cross-gating strategy is designed to fuse the self- and cross-attentive representations. We also employ a self-guided head to replace the feed forward layer in transformer block. This design implicitly emphasizes informative representations by measuring the confidence of each element.

Table~\ref{tab:stack_transformer} reports the performance of SGPA and standard TRMs on Charades-STA and ANetCaps datasets. Here, we regard both SeCo-TRM and CoSe-TRM as single block. The results show that both PA (a SGPA variant without self-guided head) and SGPA are superior to standard TRMs.

\begin{figure*}[t]
    \centering
	\subfigure[\small Transformer Block (Self Attn)]
	{\label{fig:se_trm}	\includegraphics[width=0.32\textwidth]{figures/SeTRM}}
	\subfigure[\small Transformer Block (Cross Attn)]
	{\label{fig:co_trm}	\includegraphics[width=0.32\textwidth]{figures/CoTRM}}
	\subfigure[\small Self Guided Parallel Attention (Ours)]
	{\label{fig:sgpa_trm}\includegraphics[width=0.32\textwidth]{figures/SGPA_TRM}}
	\caption{\small The structures of standard transformer blocks and self-guided parallel attention module. Top: the structure of each module; Bottom: the parallel streams of encoding visual and textual inputs. (a) The standard transformer block with self-attention; (b) The standard transformer block with cross-attention; (c) The self-guided parallel attention (SGPA) module.}
	\label{fig:transformer}
\end{figure*}

\begin{figure}[t]
    \centering
	\subfigure[\small SeCo-TRM Block]
	{
	\label{fig:seco_trm}	
	\includegraphics[trim={0mm 11mm 8mm 2mm},clip,width=0.22\textwidth]{figures/SeCoTRM}
	}
	\subfigure[\small CoSe-TRM Block]
	{
	\label{fig:cose_trm}	
	\includegraphics[trim={0mm 11mm 8mm 2mm},clip,width=0.22\textwidth]{figures/CoSeTRM}
	}
	\caption{\small The structures of SeCo-TRM and CoSe-TRM.}
	\label{fig:stack_transformer}
\end{figure}

\begin{table}[t]
    \small
	\centering
	\begin{tabular}{l | c c c}
		\specialrule{.1em}{.05em}{.05em}
		\multirow{2}{*}{Methods} & \multicolumn{3}{c}{$\text{R@}1, \text{IoU}=\mu$} \\
        & $\mu=0.3$ & $\mu=0.5$ & $\mu=0.7$ \\
        \hline
        \multicolumn{4}{c}{Charades-STA} \\
        \hline
        Se-TRM & 68.84~\scriptsize{(0.46)} & 51.92~\scriptsize{(0.54)} & 34.58~\scriptsize{(0.18)} \\
        Co-TRM & 69.03~\scriptsize{(0.49)} & 52.34~\scriptsize{(0.50)} & 35.07~\scriptsize{(0.32)} \\
        SeCo-TRM & 69.11~\scriptsize{(0.24)} & 52.63~\scriptsize{(0.49)} & 35.17~\scriptsize{(0.22)} \\
        CoSe-TRM & 69.08~\scriptsize{(0.26)} & 52.82~\scriptsize{(0.43)} & 35.09~\scriptsize{(0.50)} \\
        PA     & 69.21~\scriptsize{(0.27)} & 54.37~\scriptsize{(0.46)} & 36.22~\scriptsize{(0.49)} \\
        SGPA   & 69.47~\scriptsize{(0.32)} & 54.63~\scriptsize{(0.43)} & 36.36~\scriptsize{(0.24)} \\
        \hline
        \multicolumn{4}{c}{ActivityNet Captions} \\
        \hline
        Se-TRM & 57.64~\scriptsize{(0.38)} & 40.76~\scriptsize{(0.35)} & 25.10~\scriptsize{(0.30)} \\
        Co-TRM & 57.39~\scriptsize{(0.29)} & 40.55~\scriptsize{(0.45)} & 24.85~\scriptsize{(0.47)} \\
        SeCo-TRM & 57.47~\scriptsize{(0.38)} & 40.70~\scriptsize{(0.24)} & 25.07~\scriptsize{(0.21)} \\
        CoSe-TRM & 57.72~\scriptsize{(0.41)} & 40.85~\scriptsize{(0.17)} & 25.16~\scriptsize{(0.15)} \\
        PA     & 58.27~\scriptsize{(0.13)} & 41.59~\scriptsize{(0.24)} & 25.88~\scriptsize{(0.28)} \\
        SGPA   & 58.40~\scriptsize{(0.31)} & 41.72~\scriptsize{(0.19)} & 26.07~\scriptsize{(0.16)} \\
        \specialrule{.1em}{.05em}{.05em}
	\end{tabular}
	\caption{\small Comparison between SGPA with standard transformer blocks on Charades-STA and ANetCaps, where PA is the SGPA without self-guided head (\ie replaced by FFN) The scores in bracket denotes standard deviation.}
	\label{tab:stack_transformer}
\end{table}

\subsection{Video-Query Integration Computation}\label{apd:sec:vqi}
This section presents the detailed computation process of video-query integration (see Section 3.1.3). 

Given two inputs $\bm{X}\in\mathbb{R}^{d\times N_x}$ and $\bm{Y}\in\mathbb{R}^{d\times N_y}$, the context-query attention first computes similarities between each pair of $\bm{X}$ and $\bm{Y}$ elements as:
\begin{equation}
    \mathcal{S}=\bm{X}^{\top}\cdot\bm{W}\cdot\bm{Y}
\end{equation}
where $\bm{W}\in\mathbb{R}^{d\times d}$ and $\mathcal{S}\in\mathbb{R}^{N_x\times N_y}$. Then $X$-to-$Y$ and $Y$-to-$X$ attention weights are computed by:
\begin{equation}
\begin{aligned}
    \mathcal{A}_{\text{XY}} & = \mathcal{S}_r \cdot \bm{Y}^{\top}\in\mathbb{R}^{N_x\times d} \\
    \mathcal{A}_{\text{YX}} & = \mathcal{S}_r \cdot \mathcal{S}^{\top}_c \cdot \bm{X}^{\top}\in\mathbb{R}^{N_x\times d}
\end{aligned}
\end{equation}
where $\mathcal{S}_{r}$ and $\mathcal{S}_{c}$ are the row- and column-wise normalization of $\mathcal{S}$ by Softmax function. The final output of context-query attention is calculated as:
\begin{equation}
    \bm{X}^{Y}=\mathtt{FFN}\big([\bm{X};\mathcal{A}_{\text{XY}}^{\top};\bm{X}\odot\mathcal{A}_{\text{XY}}^{\top};\bm{X}\odot\mathcal{A}_{\text{YX}}^{\top}]\big)
\end{equation}
where $\odot$ denotes element-wise multiplication, ``$;$'' represents concatenation operation, and $\bm{X}^{Y}\in\mathbb{R}^{d\times N_x}$. In this way, the information of $\bm{Y}$ is properly fused into $\bm{X}$.

By setting $\bm{X}=\bm{\bar{V}}\in\mathbb{R}^{d\times N}$ and $\bm{Y}=\bm{\bar{Q}}\in\mathbb{R}^{d\times M}$, we can derive the query-aware video representations $\bm{V}^{Q}\in\mathbb{R}^{d\times N}$. Similarly, the video-aware query representations $\bm{Q}^{V}\in\mathbb{R}^{d\times M}$ is obtained by setting $\bm{X}=\bm{\bar{Q}}$ and $\bm{Y}=\bm{\bar{V}}$. 

Next, we encode $\bm{Q}^{V}=[\bm{q}_{0}^{V},\dots,\bm{q}_{M-1}^{V}]$ into sentence representation $\bm{q}$ with additive attention:
\begin{equation}
\begin{aligned}
    \bm{\alpha} & = \mathtt{Softmax}\big(\bm{W}_{\alpha}\cdot\bm{Q}^{V})\big) \in \mathbb{R}^{M} \\
    \bm{q} & = \sum_{i=0}^{M-1}\alpha_i\times\bm{q}_i^V \in \mathbb{R}^{d}
\end{aligned}
\end{equation}
where $\bm{W}_{\alpha}\in\mathbb{R}^{1\times d}$. The $\bm{q}$ is then concatenated with each element of $\bm{V}^{Q}$ as $\bm{H}=[\bm{h}_{1},\dots,\bm{h}_{n}]\in\mathbb{R}^{2d\times N}$, where $\bm{h}_{i}=[\bm{v}_{i}^{Q};\bm{q}]$. Finally, the query-attended visual representation is computed as
\begin{equation}
    \bm{\bar{H}}=\bm{W}_h\cdot\bm{H}+\bm{b}_h
\end{equation}
where $W_h\in\mathbb{R}^{d\times 2d}$ and $b_h\in\mathbb{R}^{d}$ denote the learnable weight and bias, and $\bm{\bar{H}}\in\mathbb{R}^{d\times N}$.

\begin{table*}[t]
    \small
    \setlength{\tabcolsep}{2.7pt}
	\centering
	\begin{tabular}{l c c c r r r r r}
		\toprule
		Dataset & Domain & $N_{\text{V}}$ (train/val/test) & $N_{\text{A}}$ (train/val/test) & $\bar{N}_{\text{A/V}}$ & $N_{\text{Vocab}}$ & $\bar{L}_{V}$ (s) & $\bar{L}_{Q}$ & $\bar{L}_{M}$ (s) \\
		\midrule
        Charades-STA & Indoors & $5,338/\text{-}/1,334$ & $12,408/\text{-}/3,720$ & $2.42$ & $1,303$ & $30.59$ & $7.22$ & $8.22$ \\
        \midrule
        ActivityNet Captions & Open & $10,009/\text{-}/4,917$ & $37,421/\text{-}/17,505$ & $3.68$ & $12,460$ & $117.61$ & $14.78$ & $36.18$ \\
        \midrule
        TACoS~\cite{Gao2017TALLTA} & \multirow{2}{*}{Cook} & \multirow{2}{*}{$75/27/25$} & $10,146/4,589/4,083$ & $148.17$ & $2,033$ & $287.14$ & $10.05$ & $5.45$ \\
        TACoS~\cite{zhang2020learning} & & & $9,790/4,436/4,001$ & $143.52$ & $1,983$ & $287.14$ & $9.42$ & $25.26$ \\
        \bottomrule
	\end{tabular}
    \caption{\small Statistics of the evaluated video grounding benchmark datasets, where $N_{\text{V}}$ is number of videos, $N_{\text{A}}$ is number of annotations, $\bar{N}_{\text{A/V}}$ denotes the average number of annotations per video, $N_{\textrm{Vocab}}$ is the vocabulary size of lowercase words, $\bar{L}_{V}$ denotes the average length of videos in seconds, $\bar{L}_{Q}$ denotes the average number of words in the sentence queries and $\bar{L}_{M}$ represents the average length of temporal moments in seconds.}
	\label{tab-data}
\end{table*}

\subsection{Categorical Reparameterization}\label{apd:sec:gumbel}
This section provides a brief introduction of the categorical reparameterization strategy used in sequence matching module (see Section 3.1.4).

Categorical reparameterization, \eg reinforce-based approaches~\cite{sutton2000policy,schulman2015gradient}, straight-through estimators~\cite{bengio2013estimating} and Gumbel-Softmax~\cite{jang2017categorical,maddison2017the}, is a strategy that enables discrete categorical variables to back-propagate in neural networks. It aims to estimate smooth gradient with a continuous relaxation for categorical variable. In this work, we use Gumbel-Softmax to approximate the sequence labels from a probability distribution. Then those labels are applied to lookup the corresponding embeddings for region representation in the sequence matching module of SeqPAN. 

Let $\bm{x}=(x_1,\dots,x_l)$ be a categorical distribution, where $l$ is the number of categories, $x_c$ is the probability score of category $c$ and $\sum_{c=1}^{l}x_c=1$. Given the \textit{i.i.d.} Gumbel noise $\bm{g}=(g_1,\dots,g_l)$ from $\mathtt{Gumbel}(0,1)$ distribution\footnote{The $\mathtt{Gumbel}(0,1)$ distribution can be sampled using inverse transform sampling by drawing $u\sim\mathtt{Uniform}(0,1)$ and computing $g=-\log(-\log(u))$~\cite{jang2017categorical}.}, the soft categorical sample can be computed as:
\begin{equation}
    \bm{y}=\mathtt{Softmax}\big((\log(\bm{x})+\bm{g})/\tau\big)
\label{eq:gumbel_softmax_appd}
\end{equation}
where $\tau>0$ is annealing temperature, and Eq.~\ref{eq:gumbel_softmax_appd} is referred as Gumbel-Softmax operation on $\bm{x}$. As $\tau\to 0^{+}$, $\bm{y}$ is equivalent to the Gumbel-Max form~\cite{gumbel1954statistical,maddison2014asampling} as:
\begin{equation}
    \bm{\hat{y}}=\mathtt{Onehot}\big(\arg\max(\log(\bm{x})+\bm{g})\big)
\label{eq:gumbel_max_appd}
\end{equation}
where $\bm{\hat{y}}$ is an unbiased sample from $\bm{x}$ and thus we can draw differentiable samples from the distribution during training. Note, when input $\bm{x}$ is unnormalized, the $\log(\cdot)$ operator in Eq.~\ref{eq:gumbel_softmax_appd} and~\ref{eq:gumbel_max_appd} shall be omitted~\cite{jang2017categorical,dong2019searching}. During inference, discrete samples can be drawn with the Gumbel-Max trick directly.

\section{Dataset and Parameter Settings}\label{apd:sec:settings}

\subsection{Dataset Statistics}\label{apd:sec:dataset}
The statistics of the evaluated benchmark datasets are summarized in Table~\ref{tab-data}. \textbf{Charades-STA} dataset consists of $6,672$ videos and $16,128$ annotations (\ie moment-query pairs) in total. \textbf{ActivityNet Captions (ANetCaps)} dataset is taken from the ActivityNet~\cite{heilbron2015activitynet}. The average duration is about $120$ seconds and each video contains $3.68$ annotations on average. \textbf{TACoS} dataset contains $127$ cooking activities videos with average duration of $4.79$ minutes, and $18,818$ annotations in total. We follow the same train/val/test split as \citet{Gao2017TALLTA}. Besides, \citet{zhang2020learning} pre-processes the TACoS dataset, hence their version is slightly different from the original version. Detailed statistics are summarized in Table~\ref{tab-data}.

\subsection{Hyper-Parameter Settings}\label{apd:sec:hyperparam}
We follow~\cite{ghosh2019excl,mun2020local,rodriguez2020proposal,zhang2020vslnet} and use 3D ConvNet pre-trained on Kinetics dataset (\ie I3D\footnote{https://github.com/deepmind/kinetics-i3d})~\cite{Carreira2017QuoVA} to extract visual features from videos. The maximal visual feature sequence lengths are set to $64$, $100$, and $256$ for Charades-STA, ActivityNet Captions, and TACoS, respectively. This setting is based on the average video lengths in the three datasets. The feature sequence length of a video will be uniformly downsampled if it is larger than the pre-set threshold, or zero-padding otherwise. For the language queries, we lowercase all the words and initialize them with GloVe~\cite{pennington2014glove} embeddings\footnote{http://nlp.stanford.edu/data/glove.840B.300d.zip}. The word embeddings and extracted visual features are fixed during training.

For other hyper-parameters, we use the same settings for all datasets. The dimension of the hidden layers is $128$; the head number in multi-head attention is $8$; the number of SGPA blocks ($N_{\text{SGPA}}$) is $2$; the annealing temperature $\tau$ of Gumbel-Softmax is $0.3$; The Dropout~\cite{srivastava2014dropout} is $0.2$. The maximal training epochs $E=100$ is used, with batch size of $16$ and early stopping tolerance of $10$ epochs. We adopt Adam~\cite{Kingma2015AdamAM} optimizer, with initial learning rate of $\beta_0=0.0001$, weight decay $0.01$, and gradient clipping $1.0$, to train the model. The learning rate decay strategy is defined as $\beta_{e}=\beta_0\times(1-\frac{e}{E})$, where $e$ denotes the $e$-th training epoch.

The SeqPAN is implemented using TensorFlow $\texttt{1.15.0}$ with CUDA $\texttt{10.0}$ and cudnn $\texttt{7.6.5}$. All the experiments are conducted on a workstation with dual NVIDIA GeForce RTX 2080Ti GPUs.

\bibliographystyle{acl_natbib}
\bibliography{acl2021}